# Collect and Connect Data Leaves to Feature Concepts: Interactive Graph Generation Toward Wellbeing

Yukio Ohsawa[1], Tomohide Maekawa[2], Hiroki Yamaguchi[2], Hiro Yoshida[1], Kaira Sekiguchi[1]
1: *The University of Tokyo, Tokyo 113-8656 Japan*   2: *Trust Architecture, Inc., Tokyo 107-0052 Japan*

**Abstract**
Feature concepts and data leaves have been invented using datasets to foster creative thoughts for creating well-being in daily life. The idea, simply put, is to attach selected and collected data leaves that are summaries of event flows to be discovered from corresponding datasets, on the target feature concept representing the well-being aimed. A graph of existing or expected datasets to be attached to a feature concept is generated semi-automatically. Rather than sheer automated generative AI, our work addresses the process of generative artificial and natural intelligence to create the basis for data use and reuse.
**Keywords**
Mobile-sensor data, multi-scale diversity, well-being, community, crowd

## 1. Introduction

Career, social, financial, physical, and community well-being. defined as the five factors of well-being [1] are interrelated: social relationships via interaction among communities affect individuals' health [2], and vice versa. A symposium in 2023 discussed the contribution of AI to these factors using various datasets [3].

In this paper, we present a tool for visualizing a graph showing the interconnectivity of social requirements toward well-being and useful datasets. Here, we focus on federating datasets to enhance well-being, positioning the physical dataset at the cross-point of social, community, and individual well-being.

## 2. Feature Concept with Data Leaves

The FC is an abstraction of the knowledge expected to be acquired from the dataset(s). An FC is represented in various ways, including text, images, or logical trees [4] corresponding to the requirements. DL is metadata that represents a latent event structure behind the dataset, corresponding to the FC for a single dataset. The DLs were selected and collected from a prepared DL base and connected to the FC. Coving up the FC by DLs, as shown in Figure 1, means envisioning data utilization to satisfy a requirement.

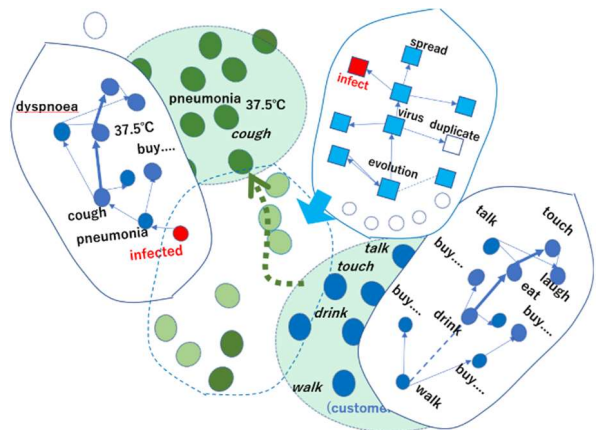

Figure 1 Feature Concept (background) and Dat Leaves (foreground leaves) attached to the FC.

## 3. Visualizing a FC/DL graph

Here, we show a tool to create a graph composed of nodes representing events or actions to satisfy a requirement, and edges implying latent causalities between them. Once the user(s) provides text $T$ about requirement(s), the tool starts the procedure below setting $G$ to null.

**Create_FC_graph ($T$)**
1: Divide $T$ into segments $\Sigma := \{S_1, S_2, \ldots S_m\}$, and create their abstracts $\alpha(S_i)$.
2: Create clusters $\{C_1, C_2, \ldots C_n\}$ of segments i.e. $C_j = \{S_k, S_{k+1}, \ldots S_{k+cj}\}$ on the semantic distance $d(S_h, S_k)$. Add node $F_j$ to $G$ and connect the $cj$ nodes in $C_j$ to $F_j$ to in $G$ via new edges.
**Collect_connect_DLs_to_G**
3: $\Delta := $ all $DL_i$, s.t. $d(DL_i, F_j) < \delta$ for each $F_j$.

---

2: Create clusters $\{Cdl_1, Cdl_2, \ldots Cdl_n\}$ of DLs in $\Delta$ s.t. $d(DL_i, F_j) < \delta$ for any $F_j$. Add node $Fdl_j$ to $G$ and connect all nodes in $Cdl_j$ to $Fdl_j$.
3: Connect each pair $(X, Y)$ to $G$ if $d(X, Y) < \delta$.

See Figure 2 for the result of $T$ on well-being in the face of COVID-19. The blue part was created by **create_FC_graph ($T$),** and the green DLs were added to $G$ in **Collect_connect_DLs_to_G**. Datasets about working conditions in schools, hospitals, and firms are added here. Datasets corresponding to DLs selected from those visualized here were then analyzed to identify factors affecting physical well-being (see Table 1).

## 4. Conclusions and Future Work

As shown in Figure 3, we are applying the presented method to enhance the well-being of Yokohama City habitats using open data [5]. The multiple colors represent FCs, DLs, and the intentions of various stakeholders in the city. Thus, we developed methods to discover new dimensions of well-being and satisfaction.

## Acknowledgments

This study was supported by JST JPMJPF2013, JSPS 23H00503, MEXT Initiative for Life Design Innovation, Cabinet Secretariat, and Social Cooperation Program on Data Federative Innovation Literacy in The University of Tokyo, and We also appreciate Yokohama Co-creation Consortium for testing the tool for FC/DL.

Table 1 The results: influence of institutes on infection expansion

| Infectious_diseases | Density (number/km2) of | | | | |
|---|---|---|---|---|---|
| | restaurants | Supermrkts | High-schls | hospitals | listed firms |
| COVID19 | **0.76** | **0.74** | 0.68 | 0.633 | **0.74** |
| Gonorrhoea | 0.53 | 0.49 | 0.48 | 0.45 | **0.64** |
| Chlamydia | 0.6 | 0.57 | 0.54 | 0.53 | **0.7** |
| Influenza | 0.38 | 0.44 | **0.51** | 0.37 | 0.4 |

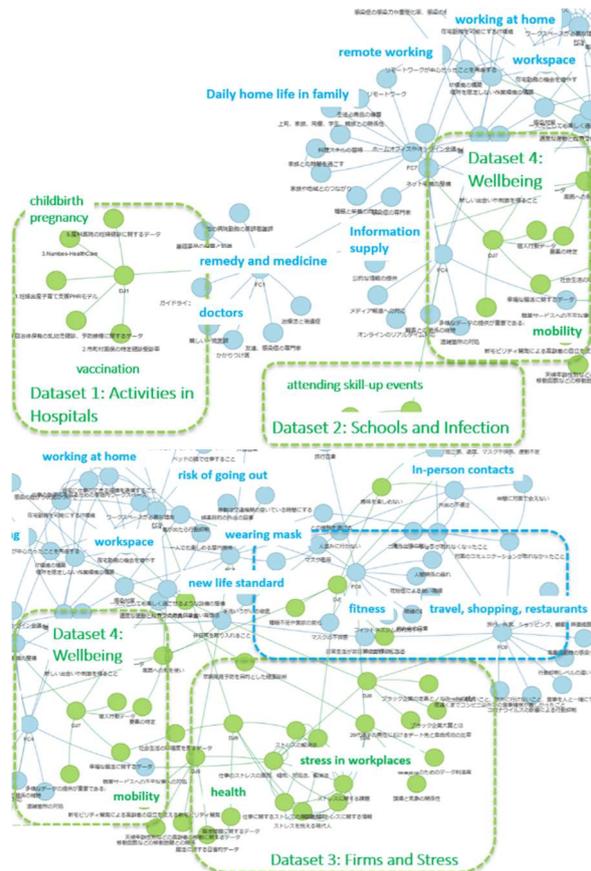

Figure 2 The left and right half of a FC/DL graph for well-beings in face of infection spreading.

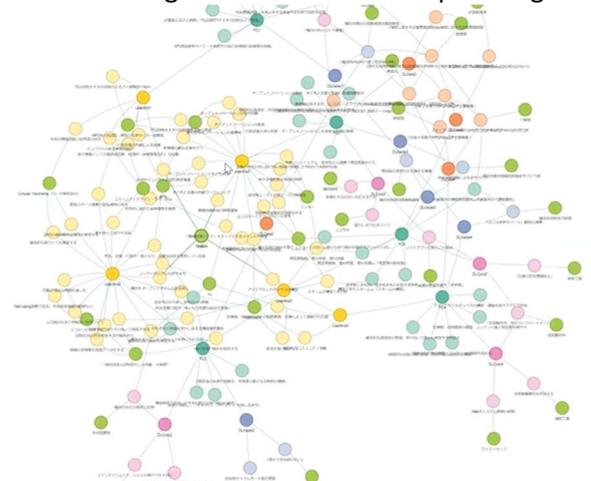

Figure 3 FC/DL for co-creation in Yokohama City.